\documentclass{llncs}

\usepackage[ruled, vlined, nofillcomment, linesnumbered]{algorithm2e}

\usepackage{cite}
\usepackage{url}
\usepackage{amsmath}
\usepackage{amssymb}
\usepackage{graphicx}
\usepackage{subfigure}
\usepackage[english]{babel}
\usepackage{booktabs}

\newcommand{\set}[1]{\left\{#1\right\}}
\newcommand{\pr}[1]{\left(#1\right)}
\newcommand{\fpr}[1]{\mathopen{}\left(#1\right)}

\newcommand{\abs}[1]{{\left|#1\right|}}

\newcommand{\enset}[2]{\left\{#1 ,\ldots , #2\right\}}

\newcommand{\naturals}{\mathbb{N}}
\newcommand{\funcdef}[3]{{#1}:{#2} \to {#3}}

\newcommand{\define}{\leftarrow}

\newcommand{\ones}[1]{o\fpr{#1}}

\newcommand{\iset}[2]{{{#1}\cdots{#2}}}
\newcommand{\freq}[1]{fr\fpr{#1}}

\newcommand{\ent}[1]{H\fpr{#1}}
\newcommand{\pemp}{p^*}

\SetKwComment{tcpas}{\{}{\}}
\SetCommentSty{textnormal}
\SetArgSty{textnormal}
\SetKw{False}{false}
\SetKw{True}{true}
\SetKw{Null}{null}
\SetKw{Output}{output}
\SetKw{AND}{and}

\newcommand{\laz}[1]{l\fpr{#1}}
\newcommand{\last}[1]{\mathit{last}\fpr{#1}}
\newcommand{\first}[1]{\mathit{first}\fpr{#1}}
\newcommand{\hideenv}[1]{}

\begin{document}
\title{Using Background Knowledge to Rank Itemsets} 
\author{Nikolaj Tatti\thanks{Nikolaj Tatti is funded by FWO Postdoctoral Mandaten.} \and Michael Mampaey\thanks{Michael Mampaey is supported by the Institute for the Promotion of Innovation through Science and Technology in Flanders (IWT-Vlaanderen).}}
\institute{Universiteit Antwerpen\\
\url{{nikolaj.tatti,michael.mampaey}@ua.ac.be}}

\maketitle

\begin{abstract}
Assessing the quality of discovered results is an important open problem in data 
mining. Such assessment is particularly vital when mining itemsets, since 
commonly many of the discovered patterns can be easily explained by background 
knowledge. The simplest approach to screen uninteresting patterns is to compare 
the observed frequency against the independence model. Since the parameters for 
the independence model are the column margins, we can view such screening as a 
way of using the column margins as background knowledge. 

In this paper we study techniques for more flexible approaches for infusing 
background knowledge. Namely, we show that we can efficiently use additional 
knowledge such as row margins, lazarus counts, and bounds of ones. We 
demonstrate that these statistics describe forms of data that occur in practice 
and have been studied in data mining. 

To infuse the information efficiently we use a maximum entropy approach. In 
its general setting, solving a maximum entropy model is infeasible, but we 
demonstrate that for our setting it can be solved in polynomial time. Experiments 
show that more sophisticated models fit the data better and that using more 
information improves the frequency prediction of itemsets.

\end{abstract}

\section{Introduction}

Discovering interesting itemsets from binary data is one of the most
studied branches in pattern mining. The most common way of defining the
interestingness of an itemset is by its frequency, the fraction of transactions in
which all items co-occur. This measure has a significant computational
advantage since frequent itemsets can be discovered using level-wise
or depth-first search strategies~\cite{agrawal:96:fast, zaki:00:scalable}. The drawback of 
frequency is that we cannot infuse any background knowledge into the ranking.
For example, if we know that items $a$ and $b$ occur often, then we should
expect that $ab$ also occurs relatively often.

Many approaches have been suggested in order to infuse background knowledge
into ranking itemsets.  The most common approach is to compare the observed
frequency against the the independence model. This
approach has the advantage that we can easily compute the estimate and that the
background knowledge is easily understandable. The downside of the independence
model is that it contains relatively little information. For example, if we
know that most of the data points contain a small number of ones, then we
should infuse this information for ranking patterns.  For a more detailed
discussion see Section~\ref{sec:related}.

Assessing the quality of patterns can be seen as a part of the
general idea where we are required to test whether a data mining
result is statistically significant with respect to some background
knowledge (see~\cite{hanhijarvi:09:tell} as an example of such a framework). 
However, the need for such assessment is especially
important in pattern mining due to two major problems: Firstly, the number of
discovered patterns is usually large, so a screening/ranking process is needed.
Secondly,
many of the discovered patterns reflect already known information, so we need to
incorporate this information such that we can remove trivial results.

Our goal in this work is to study how we can infuse background knowledge into
pattern mining efficiently. We will base our approach on building a statistical
model from the given knowledge. We set the following natural
goals for our work:
\begin{enumerate}
\item The background knowledge should be simple to understand.
\item We should be able to infer the model from the data efficiently.
\item We should be able to compute expected frequencies for itemsets efficiently.
\end{enumerate}
While these goals seem simple, they are in fact quite strict. For example,
consider modeling attribute dependencies by using Bayesian
networks. First of all, inferring the expected frequency of an itemset from the
Bayesian network is done using a message passing algorithm, and is not guaranteed
to be computationally feasible~\cite{cowell:99:probabilistic}. 
Finally, understanding parent-child relationships in a Bayesian network can be discombobulating.

In this work we will consider the following simple statistics: column margins,
row margins, number of zeroes between ones (lazarus counts), and the boundaries
of ones. We will use these statistics individually, but also consider different
combinations.  While these are simple statistics, we will show that they
describe many specific dataset structures, such as banded datasets or nested
datasets.

We will use these statistics and the maximum entropy principle to build a
global model. In a general setting inferring such a model is an infeasible
problem.  However, we will demonstrate that for our statistics inferring the
model can be done in polynomial time.  Once this model is discovered we can use
it to assess the quality of an itemset by comparing the observed frequency
against the frequency expected by the model. The more the observed frequency
differs from the expected value, the more significant is the pattern.

We should point out that while our main motivation is to assess itemsets,
the discovered background model is a true statistical global model and is useful
for other purposes, such as model selection or data generation.

\section{Statistics as Background Knowledge}

In this section we introduce several count statistics for transactional data.
We also indicate for what kinds of datasets they are useful. We begin by
presenting preliminary notations that will be used throughout the
rest of the paper.

A \emph{binary dataset} $D$ is a collection of $\abs{D}$ \emph{transactions},
binary vectors of length $N$. The set of all possible transactions is written
as $\Omega = \set{0, 1}^N$. The $i$th element of a transaction is
represented by an \emph{attribute} $a_i$, a Bernoulli random variable. We
denote the collection of all the attributes by $A = \enset{a_1}{a_N}$. An
\emph{itemset} $X = \enset{x_1}{x_L} \subseteq A$ is a subset of attributes. We
will often use the dense notation $X = \iset{x_1}{x_L}$.
Given an itemset $X$ and a binary vector $v$ of length $L$, we use the notation
$p\fpr{X = v}$ to express the probability of $p\fpr{x_1 = v_1, \ldots, x_L =
v_L}$. If $v$ contains only 1s, then we will use the notation $p\fpr{X = 1}$.

Given a binary dataset $D$ we define $q_D$ to be an \emph{empirical
distribution},
\[
    q_D\fpr{A = v} = \abs{\set{t \in D \mid t = v}} / {\abs{D}}.
\]
We define the \emph{frequency} of an itemset $X$ to be $\freq{X} = q_D\fpr{X = 1}$.
A statistic is a function $\funcdef{S}{\Omega}{\naturals}$ mapping a
transaction $t$ to a integer. All our statistics will be of form $q_D(S(A) =
k)$, that is, our background knowledge will be the fraction of transactions in
the data for which $S(t) = k$. 

\paragraph{Column margins}
The simplest of statistics one can consider are the column margins or item
probabilities. These probabilities can be used to define an independence model.
This model has been used before in other works to estimate itemset frequencies
\cite{brin:97:beyond,aggarwal:98:new}.  It has the advantage of being
computationally fast and easy to interpret. However, it is a rather
simplistic model and few datasets actually satisfy the independence assumption.
We will include the item probabilities in all of our models.

\paragraph{Row margins}
We define $\ones{A}$ to be a random variable representing 
the number of ones in a random transaction. We immediately 
see that $\ones{A}$ obtains integer values between $0$ and $N$. 
Consequently, $p(\ones{A} = k)$ is the probability of a random 
transaction having $k$ ones. Given a transaction $t\in\Omega$, 
we will also denote by $\ones{t}$ the number of ones in $t$, 
i.e.  $\ones{t}=\sum_{i=1}^{N}t_{i}$.

The use of row margins (in conjunction with column margins) has been 
proposed before by Gionis et al.~\cite{gionis:07:assessing}, 
to asses the significance of (among others) frequent itemsets.
However, their approach is different from ours (see Section~\ref{sec:related}).
It was shown that for datasets with very skewed row margin distribution, most frequent itemsets, clusterings, correlations, 
etc.\ can be explained entirely by row and column margins alone.
Supermarket basket data falls into this category, since the transactions
there typically contain only a handful of items.

\paragraph{Lazarus counts}
A \emph{lazarus event} in a transaction is defined as the occurrence of a zero within a string of ones.
This requires that a total order is specified on the attributes 
$a_{i} \in A$. For simplicity, we assume that this order is $a_{1} < \cdots <  a_{N}$.
Let $t \in \Omega$ be a transaction, then the lazarus count of $t$ is defined as
\begin{equation*}
\laz{t} = \abs{ \set{ t_{i}=0 \mid \text{there exist } a, b \text{ s.t. }  a<i<b, t_{a}= t_{b}=1}}.
\end{equation*}
Clearly, the lazarus count of a transaction $t$ ranges from $0$ to $N-2$.
If the lazarus count of $t$ is $0$, then $t$ is said to satisfy the \emph{consecutive-ones}
property.

A specific case of datasets with the consecutive-ones property are banded
datasets.  These are datasets whose rows and columns can be permuted such that
the non-zeroes form a staircase pattern.  The properties of such banded binary
matrices have been studied by Garriga et al., who presented algorithms to determine the
minimum number of bit flips it would take for a dataset to be (fully)
banded~\cite{garriga:08:banded}. A banded dataset can be characterized by the
following statistics: The major part of the transactions will have low lazarus
counts, and typically the row margins are low as well.

\paragraph{Transaction bounds}
For certain types of data it can be useful to see at which positions the ones in the transactions of a dataset
begin and end. Given a transaction $t \in \Omega$, we define the $\mathit{first}$ and $\mathit{last}$ statistics as 
\[
\first{t} = \min \{ i \mid t_{i}=1\} \text{ and } \last{t} = \max \{ i \mid t_{i}=1\}.
\]
If $t$ contains only $0$s, then we define $\first{t} = \last{t} = 0$. 
A dataset is called \emph{nested} if for each pair of rows, one is always a subset of
the other~\cite{mannila:07:nestedness}.  For such data, the rows and columns
can be permuted such that the rows have consecutive ones starting from the
first column. Thus, assuming the permutation is done, transactions in nested
datasets will have a low lazarus count and low left bound $\first{t}$.
Nestedness has been studied extensively
in the field of ecology for absence/presence data of species (see, for example,
\cite{mannila:07:nestedness} for more details).

\section{Maximum Entropy Model}
\label{sec:model}

The independence model can be seen as the simplest model that we can infer from
 binary data. This model has many nice computational properties: learning
the model is trivial and making queries on this model is a simple and fast
procedure. Moreover, a fundamental theorem shows that the independent model is the
distribution maximizing the entropy among all the distributions that have the
same frequencies for the individual attributes.

Our goal is to define a more flexible model. We require that we should be able
to infer such model efficiently and we should be able to make queries. In order to do that
we will use the maximum entropy approach. 

\subsection{Definition of the Maximum Entropy Model}

Say that we have computed a set of certain statistics from a dataset and we
wish to build a distribution, such that this distribution satisfies the same
statistics. The maximum entropy approach gives us the means to do that.

More formally, assume that we are given a function
$\funcdef{S}{\Omega}{\naturals}$ mapping a transaction $t$ to an integer value.
For example, this function can be $\ones{t}$, the number of ones in a
transaction.  Assume that we are given a dataset $D$ with $N$ attributes and
let $q_D$ be its empirical distribution.  We associate a statistic $n_k$ with $S$
for any $k = 1, \ldots, K$ to be the proportion of transactions in $D$ for
which $S(t) = k$, that is, $n_k = q_D\fpr{S(A) = k}$.

In addition to the statistics $\set{n_k}$ we always wish to use the margins
of the individual attributes, that is, the probability of an attribute having a
value of 1 in a random transaction. We denote these column margins by $m_i =
q_D\fpr{a_i = 1}$, where $i = 1, \ldots, N$.

The maximum entropy model $\pemp$ is derived from the statistics $m_i$ and $n_k$.
The distribution $\pemp$ is the unique distribution maximizing the entropy
among the distributions having the statistics $m_i$ and $n_k$. To be more precise,
we define $\mathcal{C}$ to be the set of all distributions
having the statistics $m_i$ and $n_k$,
\[
	\mathcal{C}  = \set{p \mid  p\fpr{a_i = 1} = m_i,\, p\fpr{S(A) = k} = n_k \text{ for } i = 1, \ldots, N, k = 1, \ldots, K }.
\]
The maximum entropy distribution maximizes the entropy in $\mathcal{C}$, 
\begin{equation}
\label{eq:maxentdef}
	\pemp = \underset{p}{\arg \max} \set{\ent{p} \mid p \in \mathcal{C}}.
\end{equation}
Note that $\mathcal{C}$ and consequently $\pemp$ depend on the statistics $m_i$
and $n_k$, yet we have omitted them from the notation for the sake of clarity.

Our next step is to demonstrate a classic result that the maximum entropy
distribution has a specific form.  In order to do so we begin by defining
\emph{indicator functions} $\funcdef{M_i}{\Omega}{\set{0,1}}$ such that $M_i(t)
= t_i$, that is, $M_i$ indicates whether the attribute $a_i$ has a value of $1$
in a transaction $t$. We also define $\funcdef{T_k}{\Omega}{\set{0,1}}$ such that
$T_k(t) = 1$ if and only if $S(t) = k$.

\begin{theorem}[Theorem~3.1 in \cite{csiszar:75:i-divergence}]
The maximum entropy distribution given in Eq.~\ref{eq:maxentdef} has the form
\begin{equation}
	\pemp\fpr{A = t} = \left\{
\begin{array}{ll}
u_0\prod_{i = 1}^{N}u_i^{M_i(t)}\prod_{k = 1}^{K}v_k^{T_k(t)}, &\text{ if } t \notin Z \\
0, & \text{ if } t \in Z \\
\end{array}
\right.
\label{eq:expform}
\end{equation}
for some specific set of parameters $u_i$, $v_k$ and a set $Z \subseteq \Omega$. Moreover,
a transaction $t \in Z$ if and only if $p(A = t) = 0$ for all distributions $p$ in $\mathcal{C}$.
\end{theorem}

It turns out that we can represent our maximum entropy model as a mixture
model.  Such a representation will be fruitful when we are solving and
querying the model.
To see this, let $u_i$ and $v_k$ be the parameters in Eq.~\ref{eq:expform}. We
define a distribution $q$ for which we have
\begin{equation}
	\label{eq:quidistr}
    q\fpr{A = t} = Z_q^{-1} \prod_{i = 1}^N u_i^{M_i(t)},
\end{equation}
where $Z_q$ is a normalization constant so that $q$ is a proper distribution.
Since $M_i$ depends only on $a_i$ we see that $q$ is actually an
independence distribution. This allows us to replace the parameters $u_i$ with 
more natural parameters $q_i=q(a_i=1)$. 
We should stress that $q_i$ is not equal to $m_i$.

Our next step is to consider the parameters $v_k$. First, we define a
distribution 
\begin{equation}
    \label{eq:vkident}
    r\fpr{S(A) = k} = Z_r^{-1}v_kq\fpr{S(A) = k},
\end{equation}
where $Z_r$ is a normalization constant such that $r$ is a proper distribution.
We can now express the maximum entropy model $\pemp$ using $r$ and $q$. 
By rearranging the terms in Eq.~\ref{eq:expform} we have
\begin{equation}
\label{eq:factform}
    \pemp\fpr{A = t} = \frac{v_k}{Z_r} q\fpr{A = t} = r\fpr{S(A) = k}\frac{q\fpr{A = t}}{q\fpr{S(A) = k}},
\end{equation}
where $k = S(t)$.
The right side of the equation is a mixture model. According
to this model, we first sample an integer, say $1 \leq k \leq K$, from
$r$. Once $k$ is selected we sample the actual transaction
from $q\fpr{A = t \mid S(A) = k}$.

We should point out that we have some redundancy in the definition of $r$. Namely,
we can divide each $v_k$ by $Z_r$ and consequently remove $Z_r$ from the equation.
However, keeping $Z_r$ in the equation proves to be handy later on.

\subsection{Using Multiple Statistics}
We can generalize our model by allowing multiple statistics together.  By
combining statistics we can construct more detailed models, which are based on
these relatively simple statistics.  In our case,  
building and querying the model remains polynomial in the number of attributes, when combining multiple statistics.

We distinguish two alternatives to combine count statistics.  Assume, for the
sake of exposition, we have two statistics $\funcdef{S_1}{\Omega}{\naturals}$
and $\funcdef{S_2}{\Omega}{\naturals}$, with $S_{1}(t) \leq K_{1}$ and
$S_{2}(t) \leq K_{2}$ for all $t \in \Omega$.  Then we can either consider
using the joint probabilities $q_{D}(S_{1}(A)=k_{1},S_{2}(A)=k_{2})$ for $k_{1}
\leq K_{1}$ and $k_{2} \leq K_{2}$; or, we may use the marginal
probabilities $q_{D}(S_{1}(A)=k_{1})$ and $q_{D}(S_{2}(A)=k_{2})$ separately.
The joint case can be easily reduced to the case of a single statistic by
considering $S_1 + (K_1 + 1)S_2$. 
Solving and querying the model in the marginal case can be done using 
the same techniques and the same time complexity as with the joint case.

\section{Solving the Maximum Entropy Model}
\label{sec:algorithm}

In this section we introduce an algorithm for finding the correct
distribution. We will use the classic Iterative Scaling algorithm.  For more
details of this algorithm and the proof of correctness we refer to the original
paper~\cite{darroch:72:generalized}.

\subsection{Iterative Scaling Procedure}

The generic Iterative Scaling algorithm is a framework that can be used for
discovering the maximum entropy distribution given any set of linear constraints. The idea is to
search the parameters and update them in an iterative fashion so that the
statistics we wish to constrain will converge towards the desired values.  In
our case, we update $u_i$ and $v_k$ iteratively so that the statistics of the
distribution will converge into $m_i$ and $n_k$.  The sketch of the algorithm
is given in Algorithm~\ref{alg:iterscale}.

\begin{algorithm}[htb!]
	\While{convergence has not occurred} {
		\ForEach{$i = 1, \ldots, N$} {
			\tcpas{Update $q_i$ so that $\pemp\fpr{a_i = 1} = m_i$}
			$d\define \pemp\fpr{a_i = 1}$\;
			$c \define m_i(1 - d) / ((1 - m_i)d)$\;
			$q_i \define q_ic / (1 - (1 - q_i)c)$\;
			$Z_r \define \sum_{k = 1}^K v_kq(S(A) = k)$\;
		}
		\tcpas{Update $v_k$ so that $\pemp\fpr{S(A) = k} = n_k$}
		\lForEach{$k = 1,\ldots K$}{$p_k \define \pemp\fpr{S(A) = k}$}
		\lForEach{$k = 1,\ldots K$}{$v_k \define v_kn_k / p_k$}
		$Z_r \define 1$.
	}

\caption{\textsc{IterScale} algorithm for discovering the maximum entropy
distribution satisfying the constraints $m_i$ and $n_k$. 
}
\label{alg:iterscale}
\end{algorithm}

In order to use the Iterative Scaling algorithm we need techniques for computing
the probabilities $\pemp\fpr{a_i = 1}$ and $\pemp\fpr{S(A) = k}$ (lines 3 and 7). 
The former is a special case of computing the frequency of an itemset $\pemp\fpr{X=1}$ 
and is detailed in the following subsection.
For the latter, assume that we have an algorithm, say
\textsc{ComputeProb}, which given a set of probabilities $p_1, \ldots, p_N$
will return the probabilities $p\fpr{S(A) = k}$ for each $k = 1, \ldots, K$,
where $p$ is the independence model parameterized by $p_i$,  that is, $p(a_i =
1) = p_i$.  We will now show that using only \textsc{ComputeProb} we are able
the obtain the required probabilities.

Note that the first equality in Eq.~\ref{eq:factform} implies that we
can compute $\pemp\fpr{S(A) = k}$ from $q(S(A) = k)$. To compute the latter we
call \textsc{ComputeProb} with $p_i = q_i$ (for $i = 1, \ldots, N$) as
parameters.

\subsection{Querying the Model}
We noted before that computing $\pemp\fpr{a_i = 1}$ is a special case
of computing the frequency of an itemset $\pemp\fpr{X = 1}$.
In order to do that let us write
\[
\pemp\fpr{X = 1} = \sum_{k = 1}^K \pemp\fpr{X = 1, S(A) = k}  = \sum_{k = 1}^K \frac{v_k}{Z_r}q(X = 1, S(A) = k).
\]
Let us write $y = \prod_{a_j \in X} q_j$.
Note that since $q$ is the independence model, we have $q(S(A) = k,  X = 1) = yq(S(A) = k \mid  X = 1)$. To compute $q(S(A) = k \mid  X = 1)$
we call \textsc{ComputeProb} with
parameters $p_j = q_j$ if $a_j \notin X$ and $p_j = 1$ if $a_j \in X$.

\subsection{Computing Statistics}
In order to use the Iterative Scaling algorithm we need an implementation of
\textsc{ComputeProb}, a routine that returns the probabilities of statistic $S$
with respect to the independence model. 
Since \textsc{ComputeProb} is called multiple times, the runtime consumption
is pivotal. Naturally, memory and time requirements depend on the 
statistic at hand. 
In Table~\ref{tab:statisticscomplexity} the complexities of
\textsc{ComputeProb} are listed for several statistics and joint statistics.
All statistics are computed using the same general dynamic programming idea: To
compute the statistics for items $\enset{a_1}{a_i}$ we first solve the problem
for items $\enset{a_1}{a_{i - 1}}$ and using that result we will be able to compute
efficiently the statistics for $\enset{a_1}{a_i}$. The details of these calculations are given in the appendix.

\begin{table}[htb!]
\caption{Time and memory complexity of the \textsc{ComputeProb} algorithm 
for various count statistics. $N$ is the number of attributes, and $K$ is the 
number of distinct values the statistic can assume.}
\centering
\begin{tabular}{lrr}
\toprule
Statistic & memory & time \\
\midrule
row margins (from scratch)    & $O(K)$ & $O(KN)$\\
row margins (backward method) & $O(K)$ & $O(N)$\\
lazarus counts & $O(K)$ & $O(KN)$\\
joint row margins and lazarus counts & $O(N^{2})$ & $O(K^{2}N)$ \\
joint transaction bounds & $O(N^{2})$ & $O(N^{2})$ \\
joint row margins, transaction bounds & $O(N^{3})$ & $O(KN^{2})$ \\
\bottomrule
\end{tabular}
\label{tab:statisticscomplexity}
\end{table}

\section{Experiments}
\label{sec:experiments}
In this section we present the results of experiments on synthetic and real data.
The source code of the algorithm can be found at the authors' 
website\footnote{\url{http://www.adrem.ua.ac.be/implementations/}}.

\subsection{Datasets}
The characteristics of the datasets we used are given in Table~\ref{tab:datasets}.

We created three synthetic datasets. The first one has independent items with
randomly chosen frequencies. The second dataset contains two clusters of equal size. 
In each cluster the items are independent with a frequency of 25\% and 75\% respectively. 
Hence, the row margin distribution has two peaks. In the third synthetic dataset, the items
form a Markov chain. The first item has a frequency of 50\%, and then each subsequent 
item is a noisy copy of the preceding one, that is, the item is inverted with a 25\% probability.

The real-world datasets we used are obtained from the FIMI Repository\footnote{
\url{http://fimi.cs.helsinki.fi/data/}} and \cite{myllykangas:06:dna}.
The \emph{Chess} data contains chess board descriptions. 
The \emph{DNA} dataset \cite{myllykangas:06:dna} describes 
DNA copy number amplifications. 
The \emph{Retail} dataset contains market basket data from an anonymous Belgian 
supermarket~\cite{brijs:99:using} and the \emph{Webview-1} dataset contains 
clickstream data from an e-commerce website~\cite{kohavi:00:kddcup}. 
For both datasets the rare items with a frequency lower than 
0.5\% were removed. This resulted in some empty transactions which were 
subsequently deleted.

Note that for all of the models, except the margin model, an order is needed on
the items. In this paper we resort to using the order in which the items appear 
in the data. Despite the fact that this order is not necessarily optimal, we were able to improve
over the independence model.

\begin{table}[htb!]
\caption{The datasets used in our experiments. 
Shown are the number of transactions $|D|$ and the number of attributes $N$, 
the time required to learn the various models, and the number of iterations 
until the algorithm convergences.}
\centering
\begin{tabular}{lrrrrrrrrrrrr}
\toprule
&&&&&\multicolumn{2}{c}{Margins}&&\multicolumn{2}{c}{Lazarus} && \multicolumn{2}{c}{Bounds}\\
\cmidrule{6-7}
\cmidrule{9-10}
\cmidrule{12-13}
Dataset && $|D|$ & $N$ && iter & time && iter & time && iter & time \\
\midrule
\emph{Independent}  && 100000 & 20 && 2 & 0.01 s && 2 & 0.02 s && 2 & 0.02 s \\
\emph{Clusters}  && 100000 & 20 && 2 & 0.01 s && 9 & 0.07 s && 10 & 0.07 s\\
\emph{Markov}  && 100000 & 20 && 2  & 0.01 s && 8 & 0.05 s && 12 & 0.07 s\\
\emph{Chess} && 3196 & 75 && 3 & 0.6 s && 400 & 153 s && 28 & 8 s \\
\emph{DNA} && 4590 & 391 && 8 & 313 s && 96 & 90 m && 119 & 66 m\\
\emph{Retail}  && 81998 & 221 && 4 & 26 s && 11 & 110 s && 19 & 171 s\\
\emph{Webview-1} && 52840 & 150 && 3 & 5 s && 14 & 45 s && 93 & 267 s\\
\bottomrule
\end{tabular}
\label{tab:datasets}
\end{table}

\subsection{Model Performance}

We begin by examining the log-likelihood of the datasets, given in Table~\ref{tab:loglikelihoods}. 
We train the model on the whole data and then compute its likelihood,
giving it a BIC penalty~\cite{schwarz:78:estimating},
equal to $\frac{k}{2}\log |D|$ where $k$ is the number of free parameters 
of each model, and $|D|$ is the number of transactions.
This penalty rewards models with few free parameters, while penalizing
complex ones.

Compared to the independence model, the likelihoods
are all better on all datasets with the exception of the \emph{Independent} data. 
This is expected since the \emph{Independent} data is generated 
from an independence distribution so using more advanced statistics will not improve the
BIC score.

When looking at the other two synthetic datasets, we see that the margin model
has the highest likelihood for the \emph{Clusters} data, and the lazarus model for the \emph{Markov} data.
The \emph{Clusters} data has two clusters, in both of which the items are independent.
The distribution of the row margins has two peaks, one for each cluster.
This information cannot be explained by the independence model and adding this information
improves the log-likelihood dramatically.
The items in the \emph{Markov} dataset are ordered since they form a Markov chain.
More specifically, since each item is a (noisy) copy of the previous one, we
expect the transactions to consist of only a few blocks of consecutive ones, which implies that
their lazarus count is quite low, and hence the lazarus model performs well.

\emph{Chess} is originally a categorical dataset, which has been binarized to
contain one item for each attribute-value pair.  Hence, it is a
rather dense dataset, with constant row margins, and lazarus counts and bounds
centered around a peak. That is, \emph{Chess} does not obey the
independence model and this is seen that the likelihoods of our models are
better than that of the independence model.

The likelihood of the lazarus model for the \emph{DNA} dataset, which is very close 
to being fully banded \cite{garriga:08:banded}, is significantly lower than that
of the independence model and the margin model, which indicates that
using lazarus counts are a good idea in this case. The bounds model comes
in second, also performing very well, which can again be explained by
the bandedness of the data.

Finally, \emph{Retail} and \emph{Webview-1} are sparse datasets.
The margin model performs well for both datasets. 
Therefore we can conclude that a lot of the structure of these datasets 
is captured in the row and column margins.
Interestingly, although close to the margin model, the bounds model 
is best for \emph{Webview-1}.

\begin{table*}[htb!]
\caption{BIC scores (Negative log-likelihood and BIC penalty) of the datasets for different models. 
The best (smallest) values are indicated in bold.}
\centering
\begin{tabular}{lrrrr}
\toprule
Dataset & Independent & Margins & Lazarus & Bounds \\ 
\midrule
\emph{Independent} & $\mathbf{1\,658\,486}$ & $1\,658\,630$ & $1\,658\,621$ & $1\,658\,779$ \\
\emph{Clusters} & $2\,000\,159$ & $\textbf{1\,719\,959}$ & $1\,889\,308$ & $1\,946\,942$ \\
\emph{Markov} & $2\,000\,159$ & $1\,938\,960$ & $\textbf{1\,861\,046}$ & $1\,890\,648$ \\
\emph{Chess} & $142\,054$ & $132\,921$ & $\textbf{131\,870}$ & $137\,213$ \\
\emph{DNA} & $185\,498$ & $173\,305$ & $\textbf{107\,739}$ & $109\,572$ \\
\emph{Retail} & $1\,796\,126$ & $\textbf{1\,774\,291}$ & $1\,783\,054$ & $1\,775\,588$ \\
\emph{Webview-1} & $836\,624$ & $778\,361$ & $783\,733$ & $\textbf{774\,773}$ \\
\bottomrule
\end{tabular}
\label{tab:loglikelihoods}
\end{table*}

\subsection{Frequency Estimation of Itemsets}

Next, we perform experiments on estimating the supports of a collection of itemsets.
The datasets are split in two parts, a training set and a test set. We train the models
on the training data, and use the test data to verify frequency estimates of itemsets.
We estimate the frequencies of the top $10\,000$ closed frequent itemsets in the test data 
(or all closed itemsets if there are less).

Table~\ref{tab:estimates} reports the average absolute and relative errors of 
the frequency estimates, for the independence, margins, lazarus and bounds models.
For the \emph{Independent} data, the independence model performs best, since the other models
overlearn the data. For all other datasets, except \emph{Chess}, we see that using more information
reduces both the average absolute and relative error of the frequency estimates. 
For instance, for \emph{Clusters} the average absolute error is reduced from $9.39\%$ to $0.2\%$ using
row and column margins, and likewise for \emph{Markov} the average relative error is reduced from $47.8\%$ to $21.11\%$.
The \emph{DNA}, \emph{Retail} and \emph{Webview-1} data are sparse. 
Therefore, the itemset frequencies are very low, as well as the absolute errors, 
even for the independence model. However, the relative errors are still quite high. 
In this case our models also outperform the independence model. 
For example, the relative error is reduced from $92.61\%$ to $79.77\%$ by using 
the margins model on the \emph{Webview-1} dataset. For \emph{DNA}, the average relative error
drops $5\%$ using lazarus counts.

The only exception is the \emph{Chess} dataset where the average absolute and
relative errors do not improve over the independence model. Note, however, that
this dataset is originally categorical, and contains a lot of dependencies
between the items. Interestingly enough, our models perform better than the
independence model in terms of (penalized) likelihood. This suggests that in
this case using additional information makes some itemsets interesting.

Our final experiment (given in Table~\ref{tab:itemsetloglike}) is the average improvement of log-likelihood of each 
itemset compared to the independence model: Let $f$ be the empirical frequency of itemset $X$ from the test data, and
let $p$ be the estimate given by one of the models, then the log-likelihood of $X$ is computed as
$|D| ( f \log p + (1-f) \log (1-p))$. We compute the difference between the log-likelihoods
for the independence model and the other models, and take the average over the top
$10\,000$ closed frequent itemsets.
Again, for \emph{Independent} and \emph{Chess}, the independence model performs the best.
For all the other datasets, we clearly see an improvement with respect to the independence model.
For \emph{Clusters} and \emph{Markov}, the average log-likelihood increases greatly, and is the
highest for the margin model. For \emph{DNA}, \emph{Retail} and \emph{Webview-1},
the increase in likelihood is somewhat lower. The reason for this is that both the estimates
and observed frequencies are small and close to each other.
For \emph{DNA} the average increase is highest when using lazarus counts,
while for \emph{Retail} and \emph{Webview-1} the margin model is best.

\begin{table*}[htb!]
\caption{Average absolute and relative error of the top $10\,000$ closed frequent itemsets in the test data. Lower scores are better.}
\centering
\begin{tabular}{lrrrrrrrrrr}
\toprule
&& \multicolumn{2}{c}{Independent} && \multicolumn{2}{c}{Margins}\\
\cmidrule{3-4}
\cmidrule{6-7}
Dataset && absolute & relative && absolute & relative\\
\midrule
\emph{Independent} && 0.11\% $\pm$ 0.10\% & 1.54\% $\pm$ 1.23\% && 0.11\% $\pm$ 0.10\% & 1.52\% $\pm$ 1.21\%\\
\emph{Clusters} && 9.39\% $\pm$ 0.73\% & 63.08\% $\pm$ 11.81\% && 0.20\% $\pm$ 0.11\% & 1.37\% $\pm$ 0.86\% \\
\emph{Markov} && 4.79\% $\pm$ 2.29\% & 47.80\% $\pm$ 20.86\% && 2.19\% $\pm$ 1.64\% & 21.11\% $\pm$ 13.41\% \\
\emph{Chess} && 1.81\% $\pm$ 1.35\% & 2.35\% $\pm$ 1.80\% && 1.94\% $\pm$ 1.43\% & 2.52\% $\pm$ 1.91\% \\
\emph{DNA} && 0.58\% $\pm$ 1.08\% & 85.89\% $\pm$ 30.91\% && 0.56\% $\pm$ 1.04\% & 84.27\% $\pm$ 31.76\% \\
\emph{Retail} && 0.05\% $\pm$ 0.11\% & 48.89\% $\pm$ 27.95\% && 0.04\% $\pm$ 0.09\% & 37.70\% $\pm$ 28.64\% \\
\emph{Webview-1} && 0.11\% $\pm$ 0.07\% & 92.61\% $\pm$ 17.27\% && 0.10\% $\pm$ 0.06\% & 79.77\% $\pm$ 23.94\%  \\
\midrule
&& \multicolumn{2}{c}{Lazarus} && \multicolumn{2}{c}{Bounds}\\
\cmidrule{3-4}
\cmidrule{6-7}
Dataset && absolute & relative && absolute & relative \\
\midrule
\emph{Independent} && 0.11\% $\pm$ 0.10\% & 1.54\% $\pm$ 1.23\% && 0.12\% $\pm$ 0.10\% & 1.61\% $\pm$ 1.30\% \\
\emph{Clusters} &&  7.00\% $\pm$ 1.12\% & 47.04\% $\pm$ 10.77\% && 8.21\% $\pm$ 0.93\% & 55.33\% $\pm$ 11.92\% \\
\emph{Markov} &&  2.27\% $\pm$ 1.61\% & 22.41\% $\pm$ 14.88\% && 3.36\% $\pm$ 2.24\% & 33.54\% $\pm$ 21.12\% \\
\emph{Chess} && 2.06\% $\pm$ 1.49\% & 2.68\% $\pm$ 1.99\% && 1.81\% $\pm$ 1.35\% & 2.35\% $\pm$ 1.79\% \\
\emph{DNA} && 0.45\% $\pm$ 0.75\% & 80.24\% $\pm$ 72.73\% && 0.54\% $\pm$ 0.94\% & 82.63\% $\pm$ 31.38\%\\
\emph{Retail} && 0.04\% $\pm$ 0.10\% & 42.90\% $\pm$ 27.81\% && 0.04\% $\pm$ 0.09\% & 43.71\% $\pm$ 27.27\% \\
\emph{Webview-1} &&  0.11\% $\pm$ 0.06\% & 88.35\% $\pm$ 20.87\% && 0.11\% $\pm$ 0.06\% & 90.60\% $\pm$ 18.63\%  \\
\bottomrule
\end{tabular}
\label{tab:estimates}
\end{table*}

\begin{table}[htb!]
\caption{The average improvement of itemset log-likelihoods over the independence model for the top $10\,000$ closed frequent itemsets. Higher scores are better.}
\centering
\begin{tabular}{lrrrr}
\toprule
Dataset && Margins & Lazarus & Bounds\\
\midrule
\emph{Independent} && 0.03 $\pm$ 0.12  & -0.00 $\pm$ 0.09  & -0.10 $\pm$ 0.36 \\
\emph{Clusters} && 4517.4 $\pm$ 1168.3  & 2412.6 $\pm$ 1030.1  & 1353.2 $\pm$ 660.4 \\
\emph{Markov} && 1585.9 $\pm$ 1310.5  & 1536.9 $\pm$ 1283.2  & 931.5 $\pm$ 854.1 \\
\emph{Chess} && -0.40 $\pm$ 0.52  & -0.81 $\pm$ 0.98  & 0.01 $\pm$ 0.17 \\
\emph{DNA} && 119.9 $\pm$ 235.7 &  133.1 $\pm$ 274.5 &  106.2 $\pm$ 224.8 \\
\emph{Retail} && 9.69 $\pm$ 22.49  & 4.62 $\pm$ 11.05  & 5.37 $\pm$ 24.94 \\
\emph{Webview-1} && 135.79 $\pm$ 90.32  & 64.72 $\pm$ 40.41  & 57.71 $\pm$ 62.95 \\
\bottomrule
\end{tabular}
\label{tab:itemsetloglike}
\end{table}

\section{Related Work}
\label{sec:related}
A special case of our framework greatly resembles the work done
in~\cite{gionis:07:assessing}. In that work the authors propose a procedure for
assessing the results of a data mining algorithm by sampling the datasets
having the same margins for the rows and columns as the original data. While the
goals are similar, there is a fundamental difference between the
two frameworks.
The key difference 
is that we do not differentiate individual rows.
Thus we do not know that, for example, the first row in the data has $5$ ones
but instead we know \emph{how many} rows have $5$ ones. The same key difference
can be seen between our method and Rasch models where each individual row and
column of the dataset is given its own parameter \cite{rasch:60:probabilistic}.

Our approach and the approach given in~\cite{gionis:07:assessing} complement each
other. When the results that we wish to assess do not depend on the order or identity of
transactions, it is more appropriate to use our method. An example of such data
mining algorithms is frequent itemset mining.  On the other hand, if the data is 
to be treated as a \emph{collection} of transactions, e.g.\ for segmentation,
then we should use the approach in~\cite{gionis:07:assessing}. Also, our
approach has a more theoretically sound ground since for sampling 
datasets the authors in~\cite{gionis:07:assessing} rely on MCMC techniques with no
theoretical guarantees that the actual mixing has happened.

Comparing frequencies of itemsets to estimates has been studied in
several works. The most common approach is to compare the
itemset against the independence model~\cite{brin:97:beyond,aggarwal:98:new}.
A more flexible approach has been suggested
in~\cite{jaroszewicz:04:interestingness,jaroszewicz:05:fast} where the itemset
is compared against a Bayesian network. In addition, approaches where the
maximum entropy models derived from some given known itemsets are suggested
in~\cite{meo:00:theory,tatti:08:maximum}. A common problem for these more
general approaches is that the deriving of probabilities from these models is
usually too complex. Hence, we need to resort to either
estimating the expected value by sampling, or build a local model using only the
attributes occurring in the query. In the latter case, it is shown
in~\cite{tatti:06:safe} that using only local information can distort the
frequency estimate and that the received frequencies are not consistent with
each other. Our model does not suffer from these problems
since it is a global model from which the frequency estimates can be drawn
efficiently.

\section{Conclusions}
\label{sec:conclusions}
In this paper we considered using count statistics to predict itemset
frequencies. To this end we built a maximum entropy model from which we draw
estimates for the frequency of itemsets and compare the observed value against
the estimate. We introduced efficient techniques for solving and querying the
model. Our experiments show that using these additional statistics improves the
model in terms of likelihood and in terms of predicting itemset frequencies.

\bibliographystyle{plain}
\bibliography{bibliography}
\appendix
\section{Computing Statistics}
To simplify notation, we define $A_i = \enset{a_1}{a_i}$ to
be the set of the first $i$ items.

\paragraph{Row Margins}
Our first statistic is the number of ones in a transaction $\ones{t}$.

We consider a cut version of $\ones{t}$ by defining $\ones{t; K} = \min\fpr{\ones{t}, K}$.
Hence, $p(\ones{A ; K} = k)$ is the probability of a random transaction having
$k$ ones, if $k < K$. On the other hand $p(\ones{A; K} = K) = p(\ones{A} \geq
K)$ is the probability of a random transaction having at least $K$ ones.  
We can exploit the fact that $p(\ones{A}\geq K) = 1 - \sum_{k=0}^{K-1} p(\ones{A}=k)$
to reduce computation time.
The range of $\ones{t}$ is $0,\ldots,N$, whereas the range of $\ones{A; K}$ is
$0,\ldots,K$.

Our next step is to compute the probabilities $p(\ones{A ; K} = k)$ when $p$ is
the independence model. Let us write $p_i = p(a_i = 1)$, the probability of
$a_i$ attaining the value of 1. We will first introduce a way of computing the
probabilities from scratch. In order to do that, note the following identity
\begin{equation}
\label{eq:buildones}
	p(\ones{A_i; K} = k) = p_i p(\ones{A_{i - 1}; K} = k - 1)  + (1 - p_i) p(\ones{A_{i - 1}; K} = k).
\end{equation}
This identity holds since $p$ is the independence model. Hence, to compute
$p(\ones{A; K} = k)$ we start with $p(\ones{a_1; K} = k)$, add $a_2$, $a_3$,
and so on, until we have processed all variables and reached $A$. Note that we
are computing simultaneously the probabilities for all $k$. We can perform
the computation in $O(NK)$ steps and $O(K)$ memory slots.

We can improve this further by analyzing the flow in the Iterative Scaling Algorithm.
The algorithm calls \textsc{ComputeProb} either using parameters $p_1 = q_1,
\ldots, p_N = q_N$ or $p_1 = q_1, \ldots, p_i = 1, \ldots, p_N = q_N$.

Assume that we have $q(\ones{A} = k)$ from the previous computation. These
probabilities will change only when we are updating $q_i$. To achieve more
efficient computation we will first compute $q(\ones{A} = k \mid a_i = 1)$ from
$q(\ones{A} = k)$, update $q_i$ and then update $q(\ones{A} = k)$. To do that
we can reverse the identity given in Eq.~\ref{eq:buildones} into
\begin{equation}
\label{eq:backones}
	p(\ones{A - a_i} = k) =  \frac{1}{1 - p_i} \pr{p(\ones{A} = k) - p_i p(\ones{A - a_i} = k - 1)},
\end{equation}
if $k = 1, \ldots, N$. When $k = 0$ we have
	$p(\ones{A - a_i} = 0) =  p(\ones{A} = 0)/(1 - p_i)$.
Using these identities we can take a step back and remove $a_i$ from $A$. To
compute $q(\ones{A} = k \mid a_i = 1)$ we can apply the identity in
Eq.~\ref{eq:buildones} with $p_i = 1$. Once $q_i$ is updated we can again use
Eq.~\ref{eq:buildones} to update $q(\ones{A} = k)$. All these computations can
be done in $O(N)$ time. On the other hand, in order to use Eq.~\ref{eq:backones}
we must remember the probabilities $q(\ones{A} = k)$ for \emph{all} $k$, so
consequently our memory consumption rises from $O(K)$ to $O(N)$.

\paragraph{Lazarus Events}
Our aim is to efficiently compute the probabilities $p(\laz{A}=k)$ where $p$ is 
an independence distribution. Again, let us write $p_{i} = p(a_{i}=1)$. The 
desired probabilities are computed incrementally in $N$ steps, starting from 
$p(\laz{a_{1}}=k)$, up to $p(\laz{A}=k)$. 
In order to determine the Lazarus count probabilities, we use an auxiliary statistic, 
the last occurrence $\last{t}$.

We will compute the probabilities $p(\laz{A}=k, \last{A}=j)$ and then
marginalize them to obtain $p(\laz{A}=k)$.
To simplify the following notation, we define the probability $p^{(i)}(k, j) = p(\laz{A_i} = k, \last{A_i} = j)$.

First, assume that $\last{A_i}=i$, which implies that $a_{i}=1$.
In such case the Lazarus count increases by $i - \last{A_{i - 1}} - 1$.
We can write the following identity
\begin{equation}
\label{eq:lazaruslastone}
	p^{(i)}(k, i) =  p_{i} \sum_{l=0}^{k} p^{(i-1)}(l, i + l - k - 1) ,
\end{equation}
for $k=1,\ldots,i-2$. To handle the boundary cases, for $k=0$ and $i>1$ we have
\begin{equation}
\label{eq:lazaruszero}
p^{(i)}(0, i) = p_i p^{i-1}( 0, i - 1) + p_i p^{(i-1)}( 0, 0),
\end{equation}
and finally when $k=0$ and $i=1$ we obtain
\begin{equation}
p^{(1)}(0, 1) = p(\laz{a_{1}}=0, \last{a_{1}}=1) = p_{1}.
\end{equation}

Secondly, consider the instances where $\last{A_i}<i$, in which 
case  $a_{i}$ must be equal to $0$. We obtain
$p^{(i)}(k, j) =  (1-p_{i}) p^{(i-1)}( k, j)$
for all $k=0,\ldots,i-2$ and $j=0,\ldots,i-1$. 

Using the equations above, we are able to compute $p(\laz{A}=k)$ using 
$O(N^{2})$ memory slots. Specifically, for $N(N-1)/2+2$ out of all $N^{2}-1$ 
combinations of $j$ and $k$,  is $p(\laz{A}=k,  \last{A}=j)$ potentially 
nonzero. Since in each step a quadratic number of probabilities is updated, 
$O(N^{3})$ time is needed.
However, we can reduce this to quadratic time and linear space.
First of all, note that the righthand size of Equation~\ref{eq:lazaruslastone}, being a sum 
of size linear in $k$, can be rewritten as a sum of constant size. Moreover, this
sum only uses probabilities involving $\{ a_{1}, \ldots, a_{i-1} \}$ having $a_{i}=1$,
which were computed in the previous step. Hence for $k > 1$, the probability $p^{(i)}( k, i)$ is equal to
\[
\begin{split}
p_{i} \sum_{l=0}^{k} p^{(i-1)}(l, i + l - k - 1) 
& = p_{i} p^{(i-1)}(k, i - 1) + p_i\sum_{l=0}^{k - 1} p^{(i-1)}(l, i + l - k - 1) \\
& = p_{i} p^{(i-1)}(k, i - 1) + p_{i} \frac{1-p_{i-1}}{p_{i-1}} p^{(i-1)}(k - 1, i - 1).
\end{split}
\]
For $k=1$, $p^{(i)}(1, i) = p_i p^{(i-1)}(1, i - 1) + p_i p^{(i-1)}(0, i - 2))$  is equal to
\[
p_i p^{(i-1)}(1, i - 1) + p_i \frac{1 - p_{i - 1}}{p_{i - 1}}p^{(i-1)}(0, i - 1) - p_i p^{(i-1)}( 0, 0).
\]
Finally, noting that 
\[
	p(\laz{A}=k, \last{A} \leq j) = p(\laz{A}=k, \last{A} \leq j - 1) + p^{(j)}(k, j) \prod_{i=j+1}^{N} (1-p_{i}),
\]
we can compute $p(\laz{A} = k)$ by gradually summing the second terms. We can
compute $p^{(j)}( k, j)$ in constant time using only $p^{(j-1)}(k, j - 1)$ and
$p^{(j-1)}(k, j - 1)$. Hence we can discard the terms $p^{(n)}(k, n)$, where
$n < j - 1$. Therefore, only $O(N)$ memory and $O(N^2)$ time is needed.

\paragraph{Joint Transaction Bounds}
We need to compute $p(\first{A}=i, \last{A}=j)$ for an independence 
distribution $p$, with $i,j = 0, \ldots, N$.
For the sake of brevity, let us denote $p(i,j)=p(\first{A}=i, \last{A}=j)$.
Note that $p(i,j)$ is nonzero if $i=j=0$ or $0<i\leq j$.
Hence there are $(N^{2}+N) / 2 + 1$ probabilities to compute.
We distinguish three cases
\begin{equation*}
	p(i,j) =
	\begin{cases}
		\prod_{k=1}^{N} (1 - p_{k}) & \textrm{if $i=j=0$} \\
		 p_{i} \prod_{k\neq i} (1 - p_{k}) & \textrm{if $i=j\neq0$}\\
		\prod_{k=1}^{i-1} (1 - p_{k}) p_{i} p_{j} \prod_{k=j+1}^{N} (1 - p_{k}) & \textrm{if $0<i<j$}
	\end{cases}
\end{equation*}
We can construct the $p(i,j)$ in quadratic time, by looping over $i$ and $j$, and 
maintaining the products $\prod_{k=1}^{i}(1-p_{k})$ and $\prod_{k=j}^{N}(1-p_{k})$, 
which can be updated in constant time in each iteration. 
Using these products and the individual item probabilities, 
we can construct $p(i,j)$.

\end{document}